 \pdfoutput=1

\documentclass{article}
\usepackage{spconf,amsmath,epsfig,float,subfigure,multirow,graphicx,comment}

\begin{document}\sloppy

\def\x{{\mathbf x}}
\def\L{{\cal L}}

\title{JOINT CONVOLUTIONAL NEURAL PYRAMID FOR DEPTH MAP SUPER-RESOLUTION}
%
\name{Yi Xiao$^1$, Xiang Cao$^1$, Xianyi Zhu$^1$, Renzhi Yang$^1$, Yan Zheng$^2$}
\address{
1 College of Computer Science and Electronic Engineering\\
2 College of Electric and Information Engineering\\
Hunan University\\
Changsha, China\\
}

\maketitle

\begin{abstract}
High-resolution depth map can be inferred from a low-resolution one with the guidance of an additional high-resolution texture map of the same scene. Recently, deep neural networks with large receptive fields are shown to benefit applications such as image completion. Our insight is that super resolution is similar to image completion, where only parts of the depth values are precisely known. In this paper, we present a joint convolutional neural pyramid model with large receptive fields for joint depth map super-resolution. Our model consists of three sub-networks, two convolutional neural pyramids concatenated by a normal convolutional neural network. The convolutional neural pyramids extract information from large receptive fields of the depth map and guidance map, while the convolutional neural network effectively transfers useful structures of the guidance image to the depth image. Experimental results show that our model outperforms existing state-of-the-art algorithms not only on data pairs of RGB/depth images, but also on other data pairs like color/saliency and color-scribbles/colorized images.
\end{abstract}
\begin{keywords}
deep neural network, super-resolution, convolutional pyramid
\end{keywords}
\section{Introduction}
\label{sec:intro}

Acquiring accurate, high quality and HR depth information is especially important for many applications of vision related tasks, such as 3D reconstruction, virtual reality, robot vision and 3DTV. While high quality texture images can be easy acquired by a simple color camera, depth data of HR is hard to  acquire. Generally speaking, depth acquisition methods can be divided into stereo matching based methods, laser scanning based methods and range sensing based methods.

Stereo matching based methods obtain depth information by correspondence matching and triangulating two or multiple texture images. However, the performance of these methods are dramatically affected by the occlusion and distributions of textures~\cite{Scharstein2002A}. Also, high resolution results require  high computational costs. Laser scanning based methods can acquire high quality depth maps, but its slice-by-slice scanning process is rather time-consuming and infeasible in dynamic scenes. In contrast, range sensing based methods which use depth sensors, such as time-of-fight(TOF) camera and Microsoft Kinect, can be used in a dynamic environment. However, the acquired depth maps are of low resolution (512$\times$424 for Kinect 2.0).


In order to solve this problem, researchers have made a lot of efforts to improve the resolution of LR depth maps. Joint or guided image filtering methods~\cite{Kopf2007Joint,He2010Guided}, which take an extra HR texture image as a reference or guidance, have achieved great success in recent years.
The main idea of joint filtering is to transfer the structural information in the HR reference image to the up-sampled depth image, so that the missing information of the depth image are restored as much as possible. However, it is ambiguous to determine which parts should be preserved and which parts should be smoothed. Ham \emph{et al} took into account the structures in both target and guidance image instead of unilaterally transferring the structures of RGB image to the depth image~\cite{Ham2015Robust}.
However, the hand-crafted objective functions used in~\cite{Ham2015Robust} may not reflect natural image priors well. Recently, joint deep convolutional neural networks (JDCNNs)~\cite{Hui2016Depth,Li2016Deep} were proposed by combining the joint filtering and deep convolutional neural networks (CNNs). They showed better performances than classical methods in depth map super resolution.

\begin{figure}[h]
\centering
\subfigure[Joint Convolutional Neural Pyramid (JCNP)]{
\includegraphics[width=0.45\textwidth]{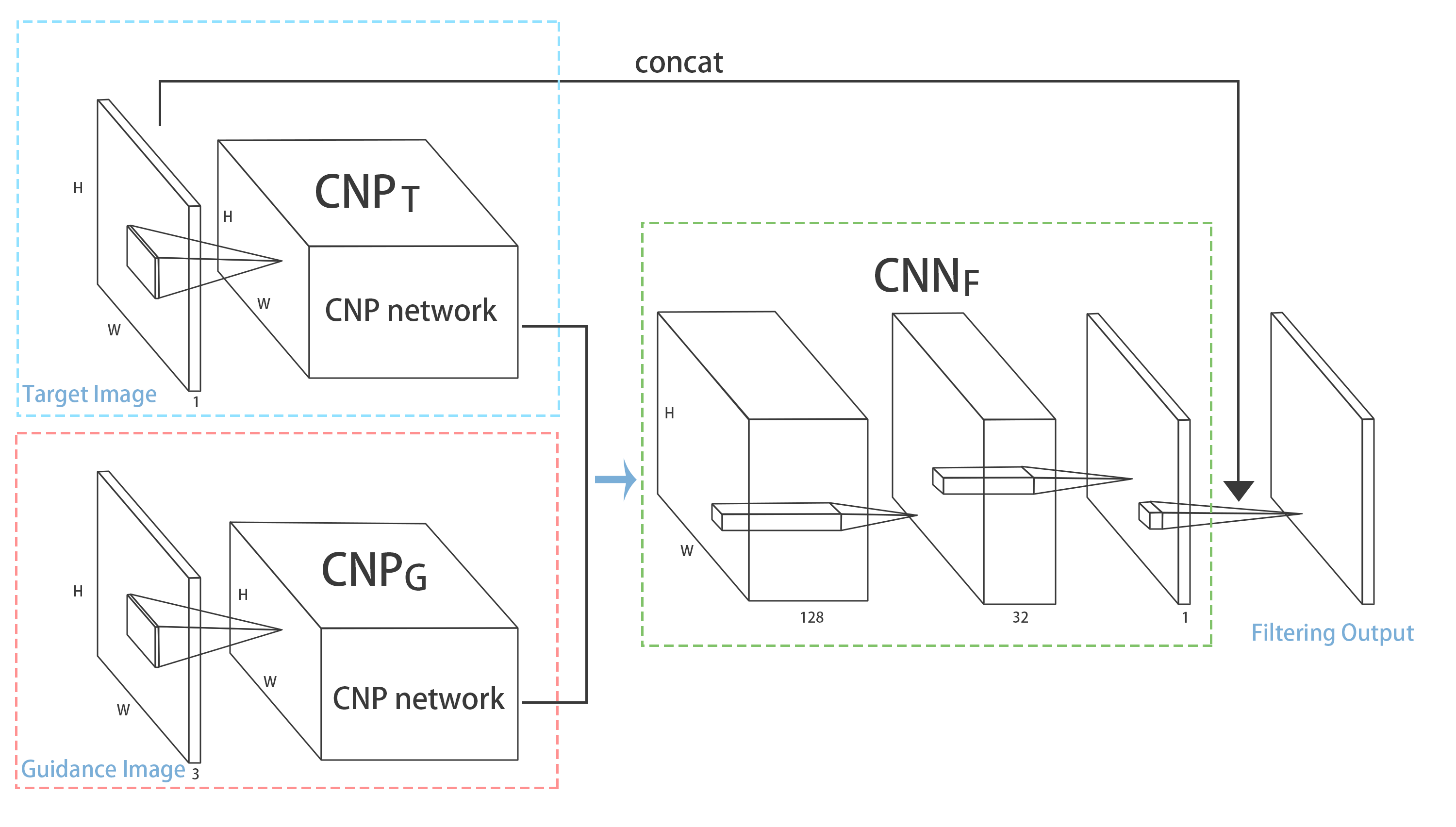}}\\
\subfigure[Convolutional Neural Pyramid (CNP)]{
\includegraphics[width=0.45\textwidth]{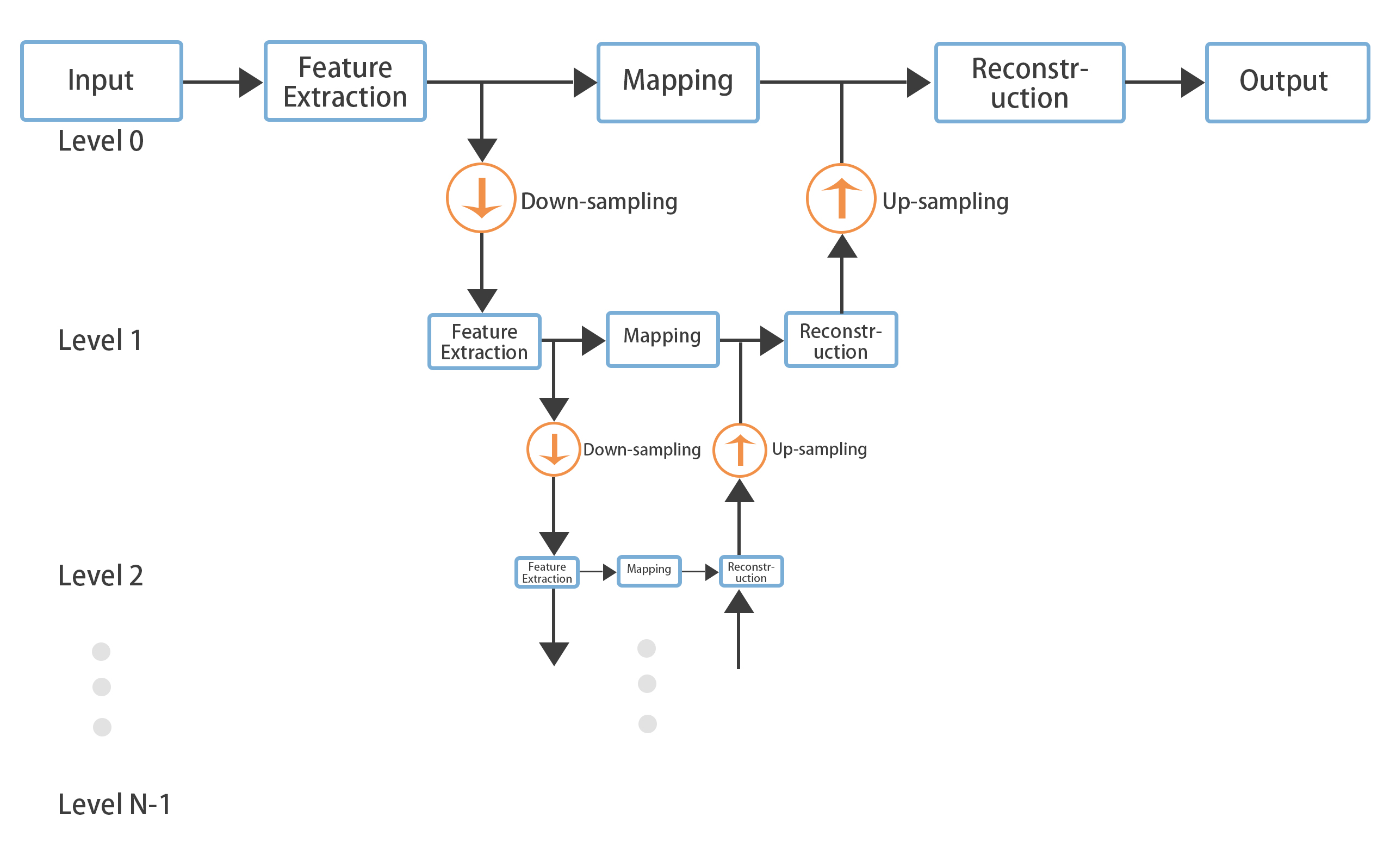}}
\caption{The network architecture of our model. (a) Joint pyramid network model. Our model consists of three sub-network:CNP$_T$, CNP$_G$ and CNN$_F$. CNP$_T$ and CNP$_G$ extract feature maps from the target and guidance images, respectively. CNN$_F$ concatenate the feature maps from CNP$_G$ and CNP$_T$ and reconstruct the desired output. (b) Convolution neural pyramid. Each pyramid level includes feature extracting part, mapping part and reconstructing part. The CNP can enlarge the receptive fields without sacrificing computational efficiency.}
\label{Figure-Overview}
\end{figure}

 As pointed out by Shen \emph{et al}, large receipt fields in deep CNN model can significantly benefit applications such as noise suppression and image completion~\cite{DBLP:journals/corr/ShenCTJ17}.
 Our insight is that super resolution is similar to noise suppression and image completion, where only parts of the depth values are precisely known. Therefore, increasing the receipt fields in the JDCNN model can also improve the performance of the network.
In this paper, we propose a novel model called joint convolutional neural pyramid (JCNP), which enables large receipt fields without significantly increasing the computational costs and memory costs. Our JCNP model main consists of three sub-networks, where two convolutional neural pyramids (CNPs) are concatenated by a CNN. The two CNPs are designed to extract information from large receptive fields, while the CNN model is designed to transfer the useful structures of the guidance images to the up-sampled target image. The flowchart of our JCNP model is shown in Figure~\ref{Figure-Overview}. To verify the performance of our model, we tested our model with several types of data pairs, including RGB/depth images, color/saliency images, and color-scribbles/colorized images. Experimental results on several benchmark data sets show that our method outperforms the existing state-of-the-art algorithms.

\section{Back Ground And Related Work}
Generally speaking, depth map SR methods can be classified into local-based, global-based and CNN-based methods.

\subsection{Local-based Methods}
In local-based depth map SR methods, each value in the SR image is given by the weighted average of its neighboring pixels in the LR depth map and guidance image~\cite{Kopf2007Joint,Yang2007Spatial,He2010Guided,Liu2013Joint,Lu2015Sparse}.
Yang \emph{et al} proposed to use a HR color map as a reference to iteratively refine LR depth maps~\cite{Yang2007Spatial}. Kopf \emph{et al} proposed a joint bilateral upsampling (JBU) procedure that considered both the depth map smoothness and the color map similarity~\cite{Kopf2007Joint}, but it may cause unwanted gradient reversal artifacts. To solve this problem, He \emph{et al} used local linear transform of the guidance image to product filtering output~\cite{He2010Guided}. Liu \emph{et al} proposed utilizing geodesic distances to upsample LR depth map with a registered HR color image~\cite{Liu2013Joint}. Lu \emph{et al} used the relationship between image segmentation boundaries and depth boundaries to reconstruct HR depth map~\cite{Lu2015Sparse}.

\subsection{Global-based Methods}
In contrast, global-based depth map SR methods~\cite{Diebel2005An,Park2011High,Aodha2012Patch,Ferstl2013Image,Yang2014Color,Ham2015Robust,Lei2017Depth} restore the SR image by solving an optimizing problem with certain regression terms.
Diebel \emph{et al} combined LR depth maps with registered HR camera maps to reconstruct HR depth maps, which may cause over-smooth depth map~\cite{Diebel2005An}.
To maintain sharp depth boundaries and prevent depth bleeding during propagation, Park \emph{et al} combined several weighting factors together with nonlocal means filtering~\cite{Park2011High}. Given a database of training patches~\cite{Aodha2012Patch}, Aodha \emph{et al} inferred the HR map from a single LR map. Ferstl \emph{et al} especially formulated SR as a global energy optimization problem using Total Generalized Variation (TGV) regularization~\cite{Ferstl2013Image}.
Yang \emph{et al} proposed an adaptive color-guided autoregressive model to construct a unified depth recovery framework~\cite{Yang2014Color}. Ham \emph{et al} fused appropriate structures of static and dynamic color maps to reconstruct HR depth maps~\cite{Ham2015Robust}. Lei \emph{et al} considered view synthesis quality to reconstruct SR depth maps~\cite{Lei2017Depth}.

\subsection{CNN-based Methods}

Convolutional neural networks (CNNs) have achieved great success in high-level computer vision~\cite{Krizhevsky2012ImageNet}. They were used to solve some low-level vision, such image SR~\cite{Dong2014Learning,Hui2016Depth,Li2016Deep,Ren2017Image}. Existing CNN-based methods take either one or two image as inputs. Dong \emph{et al} proposed an end-to-end SR convolution neural network (SRCNN) for single image SR~\cite{Dong2014Learning}. Ren \emph{et al} fused different individual networks to construct a super resolution system~\cite{Ren2017Image}. Joint deep convolutional neural networks~\cite{Hui2016Depth,Li2016Deep}, taking the depth image and guidance images as inputs, were proposed to simulate a joint filter. Hui \emph{et al} proposed a multi-scale guided convolutional network (MSG-Net) for depth map super resolution. Li \emph{et al} proposed a deep convolution network to perform joint filtering~\cite{Li2016Deep}. The model consists of three sub-networks. The first two CNNs extract informative features from both target and guidance images in parallel. The third CNN then concatenates the feature responses to selectively transfer common structures and re-construct the filtered output~\cite{Li2016Deep}. In contrast, our model uses two convolution neural pyramid~\cite{DBLP:journals/corr/ShenCTJ17} with large receptive fields as feature extractors.  In addition, we concatenate the target image at the end of our model to improve the flow of information and gradients throughout the network.

\section{Proposed Method}

\subsection{Overview}
As illustrated in Figure~\ref{Figure-Overview}(a), our joint convolutional neural pyramid (JCNP) is composed of three sub-networks, two convolutional neural pyramids (CNP$_T$, CNP$_G$) followed by a normal CNN (CNN$_F$). CNP$_T$ and CNP$_G$ concurrently extract the informative features from large receptive fields in the target and guidance images. Their output feature maps are then concatenated and feeded to CNN$_F$, which transfers useful structures from the guidance image to the target image. In order to improve the flow of information and gradients throughout the network~\cite{Huang2016Densely}, the target image is concatenated with the output of CNN$_F$ and then convolved to produce the final SR image.

\subsection{Convolutional Neural Pyramid}
Convolution neural network with large receptive fields are essential for many low-level vision tasks such as image SR and suppressing noise.
Pyramid structure can greatly enlarge the receptive fields without sacrificing computation efficiency. We adopt the CNP model~\cite{DBLP:journals/corr/ShenCTJ17} in our joint framework to extract the informative features from large receptive fields.
Each CNP level includes feature extraction/representation part, non-linear mapping part and reconstruction part, as illustrated in Figure~\ref{Figure-Overview}(b). In the following, we will introduce each part in more detail.


\textbf{Feature Extraction Part:}
The feature extraction part in level $L_i$, consisting of two convolution-PReLU layers with kernel size \emph{3$\times$3}, takes the down-sampled feature maps from the feature extraction part in $L_{i-1}$ (expect $L_0$, whose input is the target image or the guidance image) and outputs 64-channel feature maps. The output feature maps are feeded to not only the non-linear mapping part in the same level, but also to the feature extraction part in level $L_{i+1}$. Since the feature extraction in higher levels receives recursively down-sampled feature maps, it can extract information from larger recept fields with a fixed kernel size. The down-sampling operation is performed by a max pooling layer.


\textbf{Non-linear Mapping Part:}
The non-linear mapping part is designed to contain three convolution layers with kernel size \emph{3$\times$3}. The first and second convolution layers output 16-channel feature maps; third layer outputs 64-channel feature maps. And PReLU rectification is applied after each layer.

\textbf{Reconstruction Part:}
The reconstruction part in level $L_i$ fuses not only the feature maps from the nonlinear mapping part in the same level, but also the up-sampled feature maps from the reconstruction part in level $L_{i+1}$
The up-sampling operation is performed by a deconvolution layer of stride 2. Since features from level $L_{i+1}$ with larger recept fields are recursively fused to level $L_{i}$, the final reconstruction part in level $L_0$ will obtain features of a large recept field.


\subsection{Joint SR Network}
The features extracted from the target image and the guidance image by CNP$_T$ and CNP$_G$ should be fused appropriately to increase the resolution of the target image. To achieve this task, a sub-network CNN$_F$ is designed to concatenate the output feature maps of CNP$_T$ and CNP$_G$ (128 channels in total).
The CNN$_F$ consists of three convolution layers of kernel size \emph{3$\times$3}. The first convolution layer outputs 32-channel feature maps; the second layer outputs a 1-channel map; third layer outputs the final SR image.
A PReLU layer is applied after the second convolution layer. In order to better guide the training, the target image is concatenated as the input of the third convolution layer, which is shown to improve the performance in the experiments.

\subsection{Loss Function}
 We train our network by minimizing the mean squared error (MSE) for $N$ training samples as
\begin{equation*}
MSE = \frac{1}{N} \sum_{} \left\| I^{gt}-\phi(I^{G},I^{T}) \right \|^{2},
\end{equation*}
where I$^{gt}$, I$^{G}$, I$^{T}$ denote ground truth depth image, guidance image and target image, respectively. And $\phi$ denotes our joint convolutional neural pyramid model.

\section{EXPERIMENTS AND APPLICATIONS}

\subsection{Training Settings}
We implement our network in Tensorflow on an NVIDIA GeForce GTX 1080Ti graphics card. We collect 1449 RGB/D image pairs from NYU data set~\cite{Silberman2012Indoor}, 1000 RGB/D image pairs for training and 449 RGB/D image pairs for testing. We augment the training data by clip, rotation and mirror and generate 180,000 training patch pairs of size \emph{128$\times$128}. The batch size is set to 36. The network is trained with 200,000 steps and the initial learning rate is 1e-3, which decays to 0.8 times per 10000 steps. We train our network for joint image SR. We get the low-resolution target image from a ground-truth image using nearest-neighbor down-sampling. When our JCNP model is trained with the augmented RGB/D data pairs, it can be directly applied to several different joint image SR tasks, including depth map SR, chromaticity map SR and saliency map SR. In experiments, we provide the average root mean squared errors (RMSEs) of test data set, except for the visual results, to evaluate the results.

\subsection{Pyramid Levels vs. Performance}
Since the number of pyramid levels $N$ in our JCNP model will affect the performance of our network, we test the impact of $N$ on joint depth map SR using the data set of Lu~\cite{Lu2014Depth}. As shown in Table~\ref{tab:memory and time}, our method shows the best results when $N=2$. Therefore, we set $N=2$ in all our experiments. We also compare the recept field, runtime, memory consumption, number of parameters and performance of our JCNP model with those of the single-level CNN model~\cite{Li2016Deep} (similar to level 0 of our model) as shown in Table~\ref{tab:memory and time}. To increase the recept field of the model in~\cite{Lu2014Depth}, we increase the number of convolution layers. As shown in Table~\ref{tab:memory and time}, our JCNP model can enlarge the receptive fields and produce better results without significantly sacrificing computational efficiency.
\begin{table*}[t]
\centering
\caption{Time(second) of each 100 steps with batch size 1,  network parameters(million), RMSEs and memory consumption(MB) analysis in terms of different sizes of receptive fields. The RMSE of 58 Layers is missing, because our GPU memory does not allow 58 layers in training. The training batch size is 36.} \label{tab:memory and time}
\begin{tabular}{|c|c|c|c|c|c|c|c|c|c|c|}
  \hline
  {} & \multicolumn{5}{c|}{Our CPN} & \multicolumn{5}{c|}{Single-Level CNN}
  \\
  \hline
  RF  & Levels & Time & Mem.  & Paras. & RMSE  & Layers & Time & Mem.   & Paras. & RMSE   \\ \hline
  17  & 0      & 1.47 & 38.50 & 0.15   & 7.42  & 8      & 1.47 & 38.50  & 0.15   & 7.42   \\ \hline
  26  & 1      & 2.27 & 63.50 & 0.41   & 6.64  & 13     & 3.37 & 84.50  & 0.74   & 7.37   \\ \hline
  56  & 2      & 2.90 & 69.75 & 0.67   & 6.28  & 28     & 8.13 & 204.50 & 1.85   & 7.05   \\ \hline
  116 & 3      & 3.42 & 71.31 & 0.92   & 6.39  & 58     & 17.68& 444.50 & 4.07   & $-$   \\ \hline
\end{tabular}
\end{table*}

\begin{figure*}
\centering
\subfigure[RGB]{
\begin{minipage}[b]{0.1\textwidth}
\includegraphics[width=\textwidth]{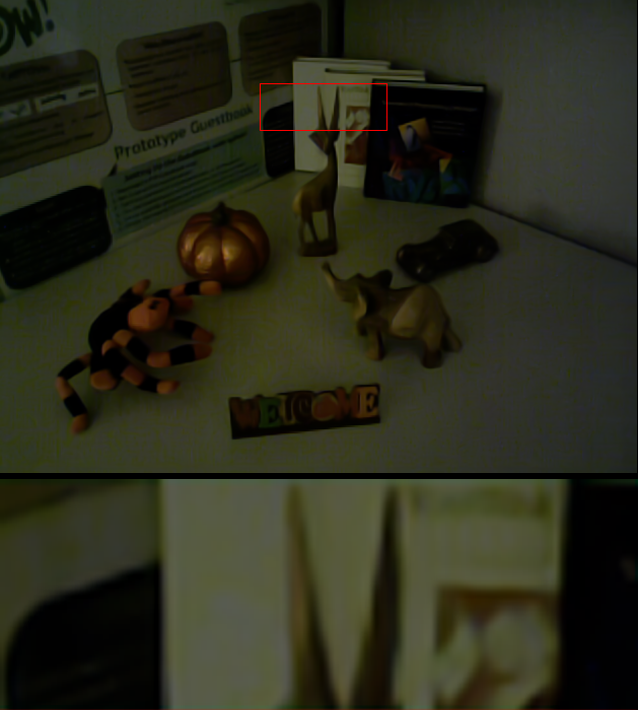} \\
\includegraphics[width=\textwidth]{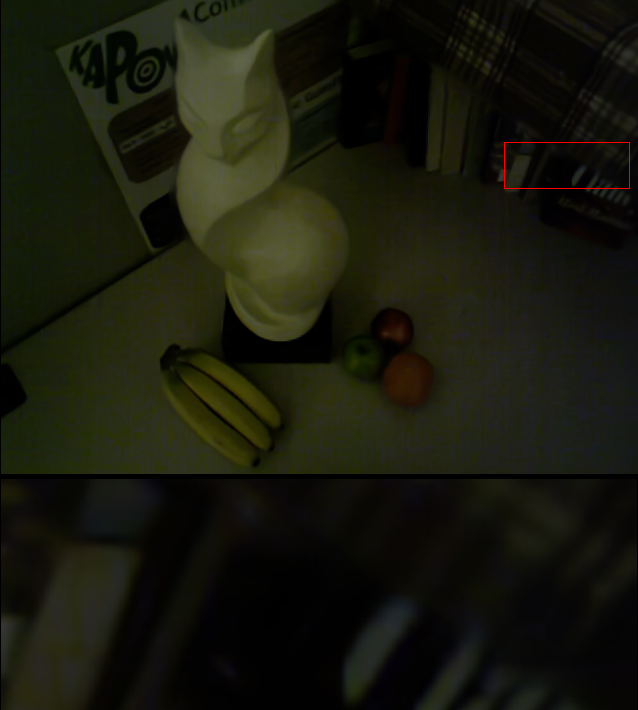} \\
\includegraphics[width=\textwidth]{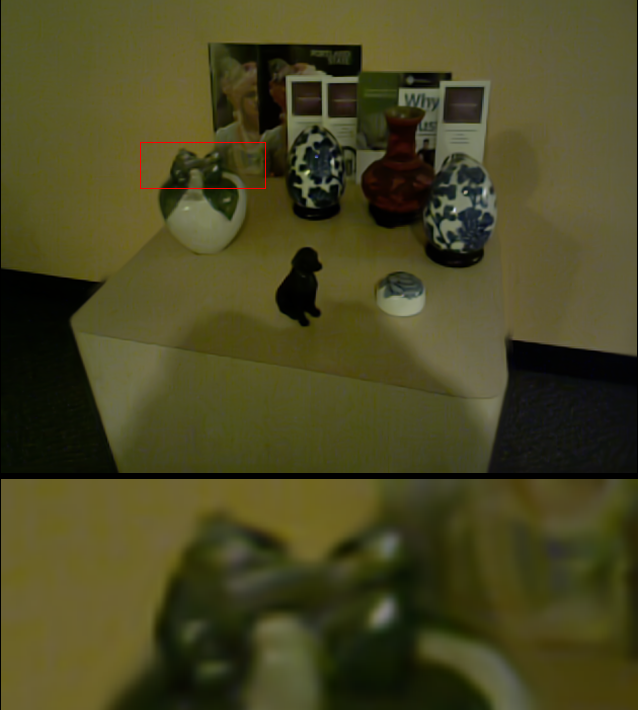}
\end{minipage}
}
\subfigure[GT]{
\begin{minipage}[b]{0.1\textwidth}
\includegraphics[width=\textwidth]{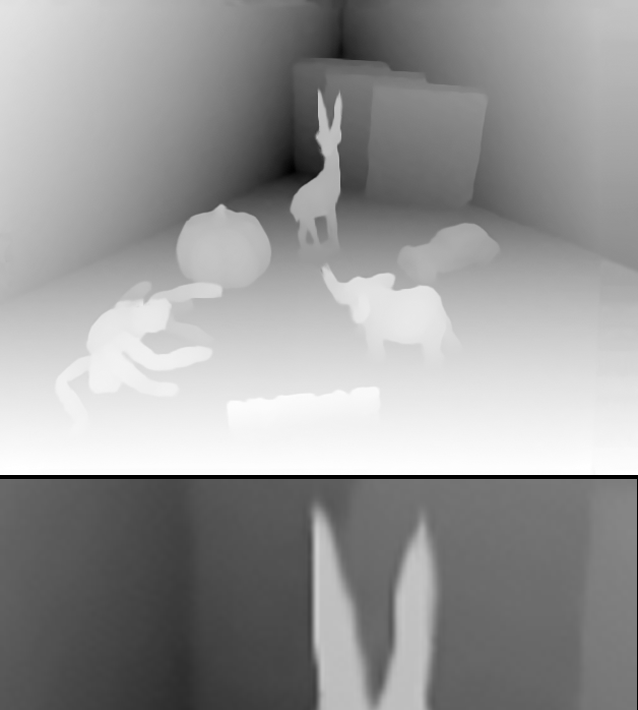}\\
\includegraphics[width=\textwidth]{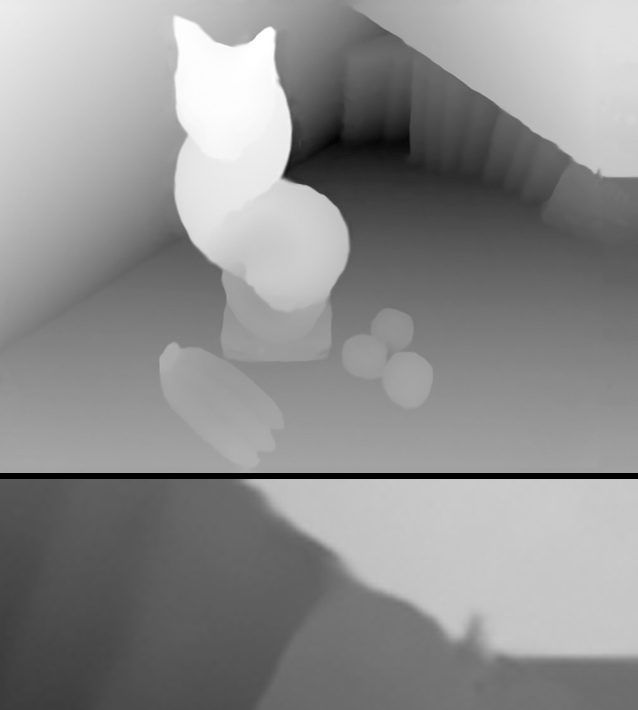}\\
\includegraphics[width=\textwidth]{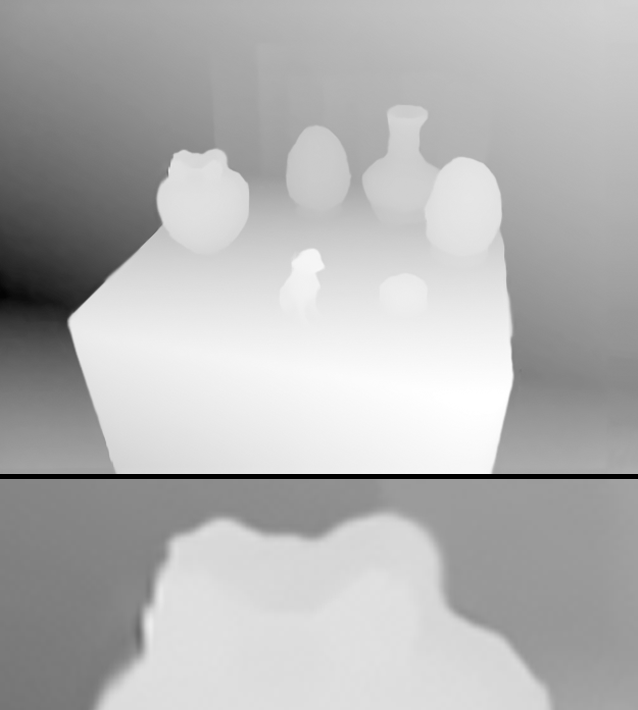}
\end{minipage}
}
\subfigure[Bicubic]{
\begin{minipage}[b]{0.1\textwidth}
\includegraphics[width=\textwidth]{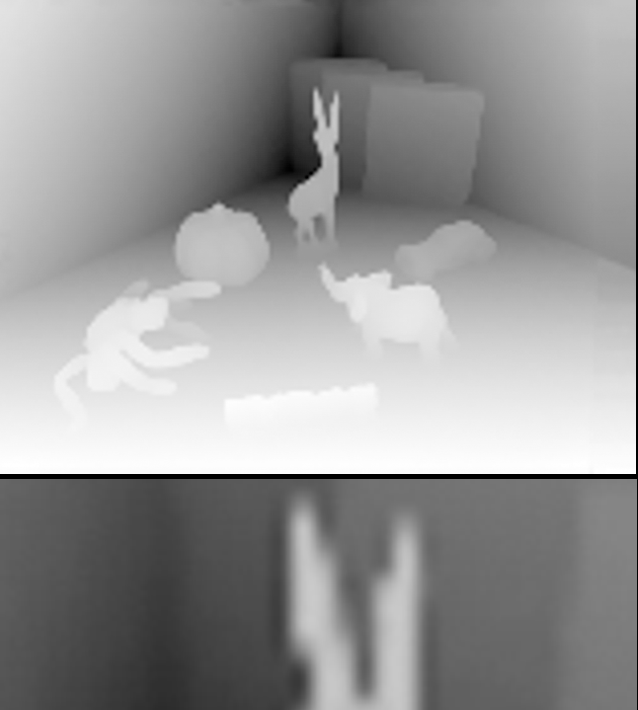}\\
\includegraphics[width=\textwidth]{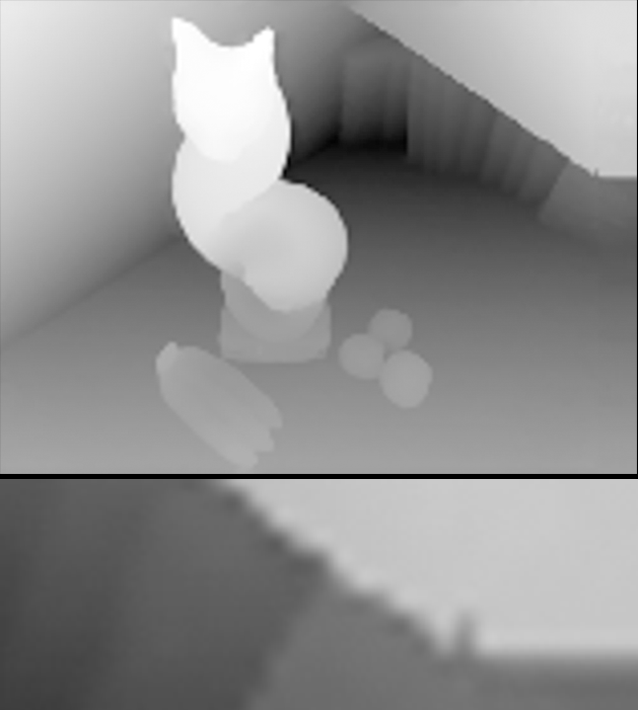}\\
\includegraphics[width=\textwidth]{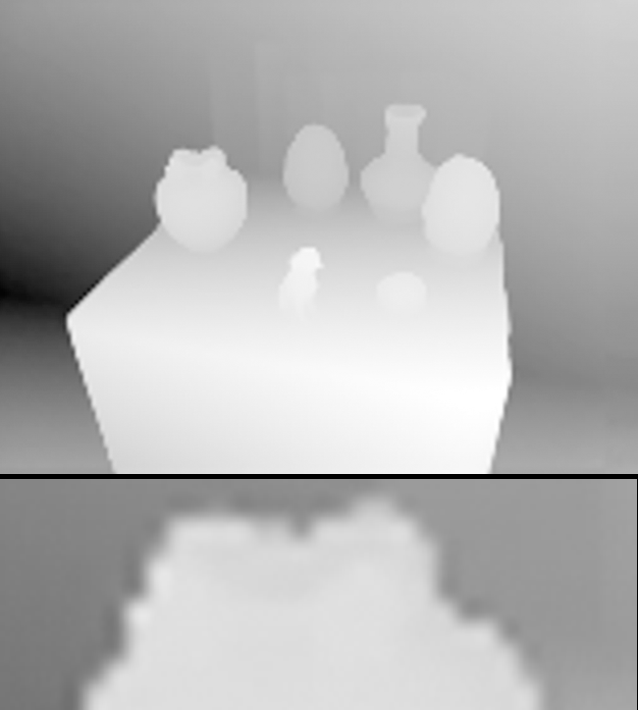}
\end{minipage}
}
\subfigure[He~\cite{He2010Guided}]{
\begin{minipage}[b]{0.1\textwidth}
\includegraphics[width=\textwidth]{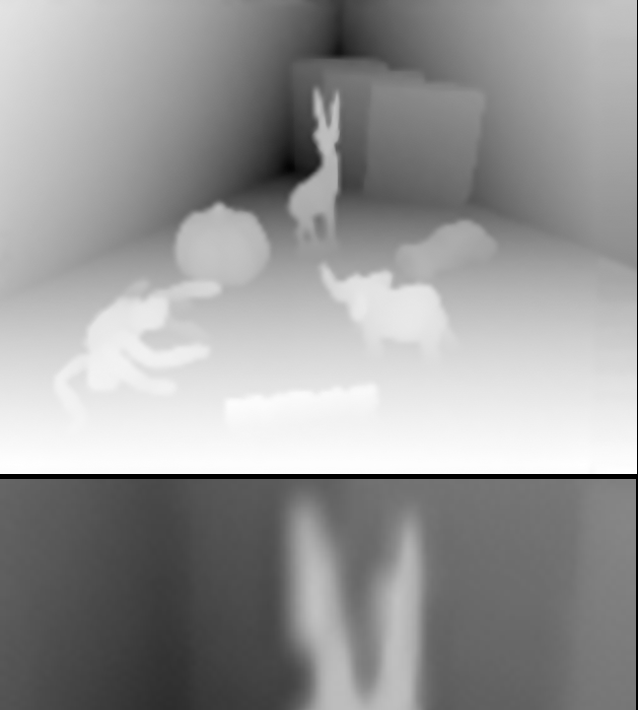}\\
\includegraphics[width=\textwidth]{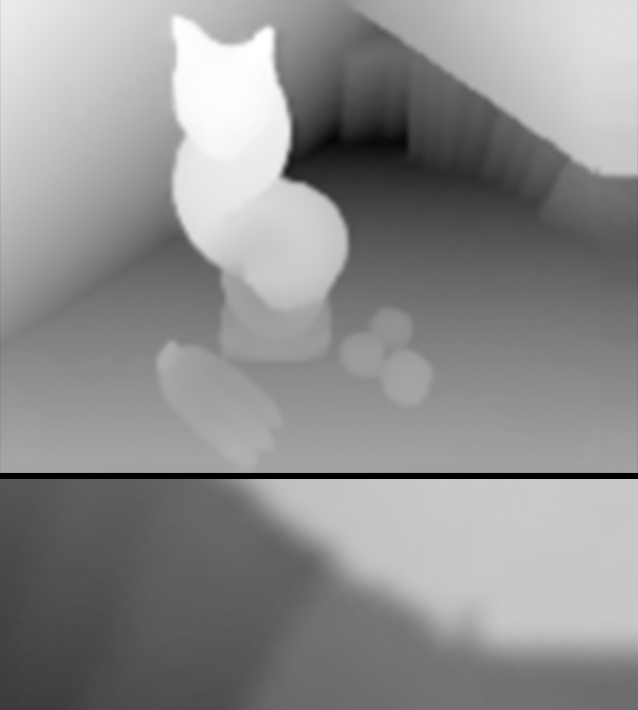}\\
\includegraphics[width=\textwidth]{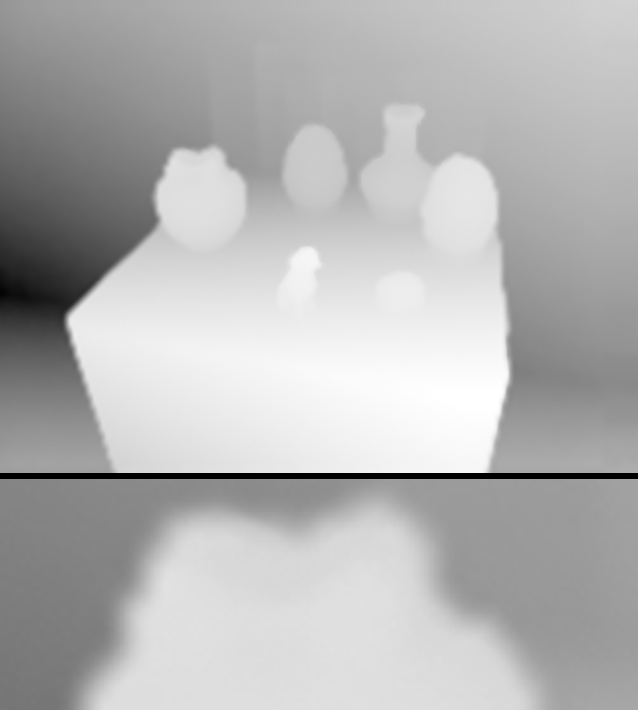}
\end{minipage}
}
\subfigure[Li~\cite{Li2016Deep}]{
\begin{minipage}[b]{0.1\textwidth}
\includegraphics[width=\textwidth]{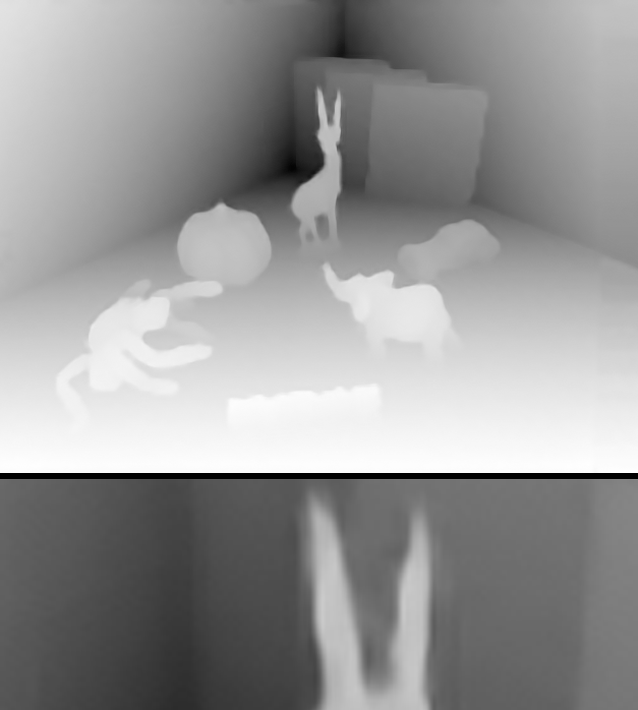}\\
\includegraphics[width=\textwidth]{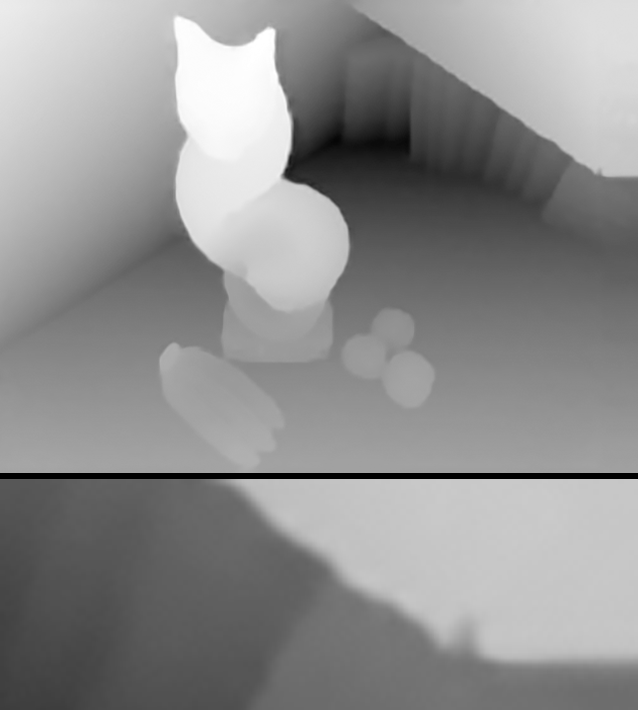}\\
\includegraphics[width=\textwidth]{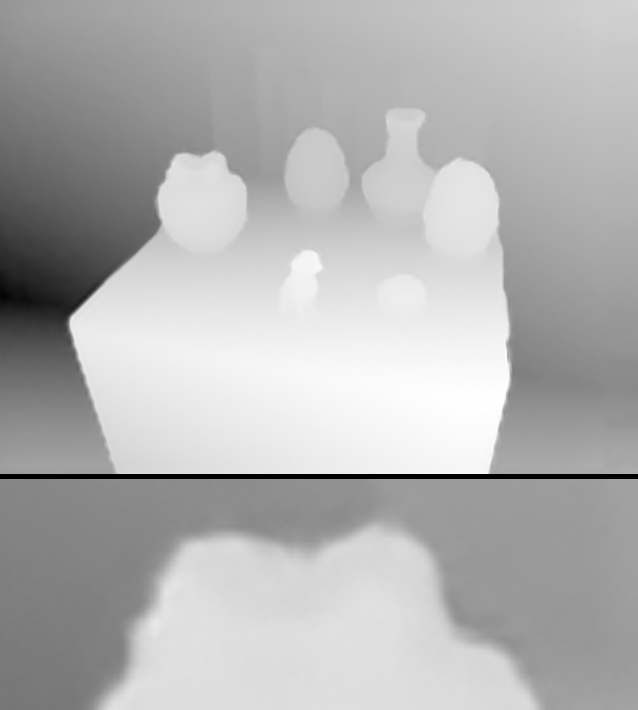}
\end{minipage}
}
\subfigure[JBU~\cite{Kopf2007Joint}]{
\begin{minipage}[b]{0.1\textwidth}
\includegraphics[width=\textwidth]{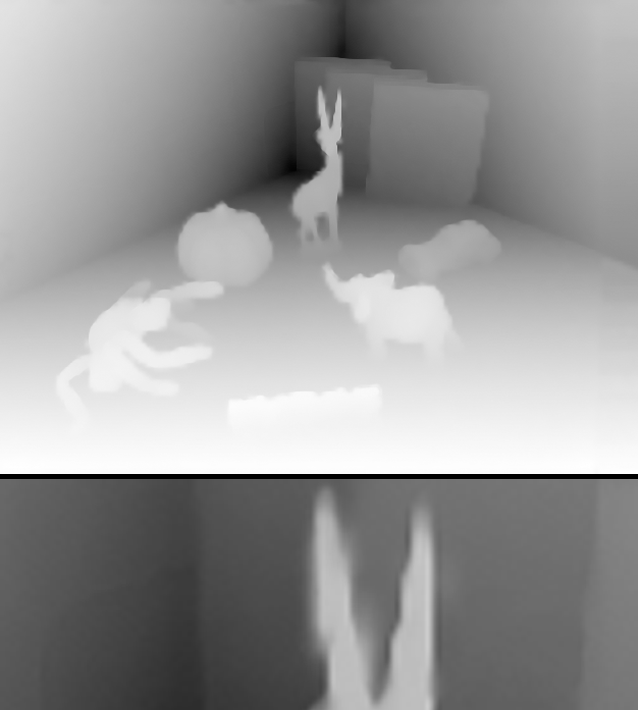}\\
\includegraphics[width=\textwidth]{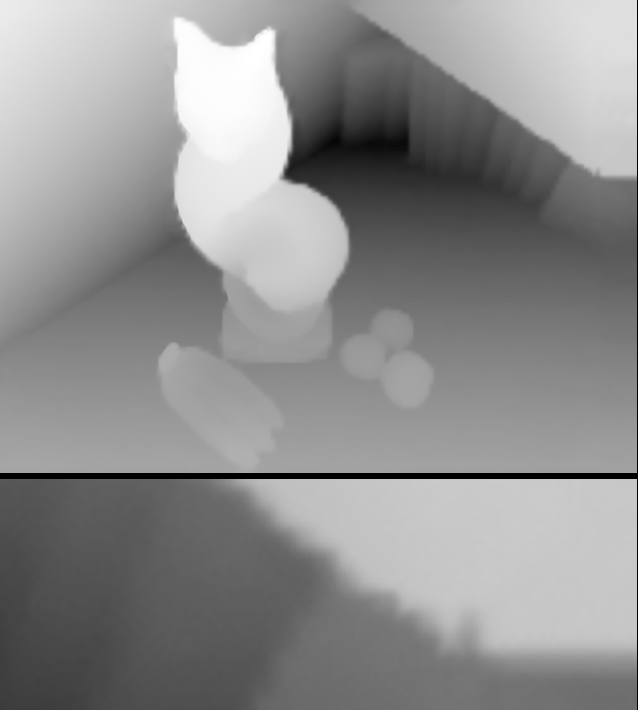}\\
\includegraphics[width=\textwidth]{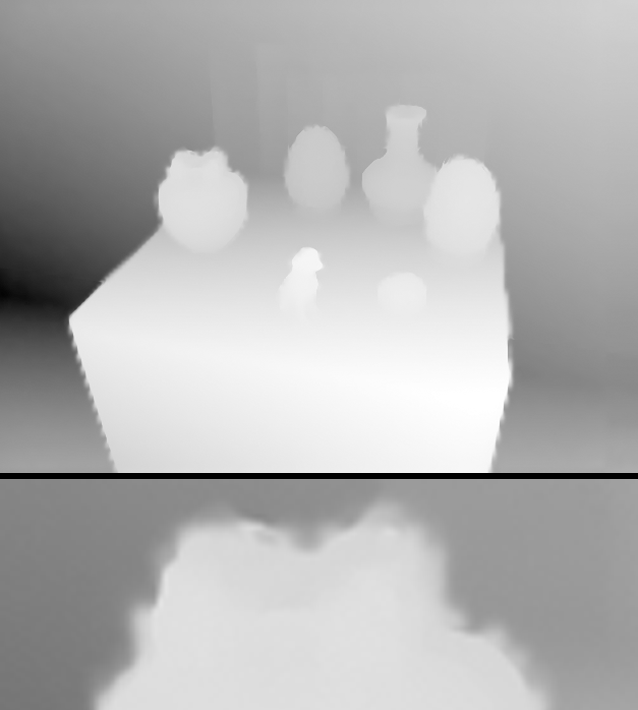}
\end{minipage}
}
\subfigure[Ham~\cite{Ham2015Robust}]{
\begin{minipage}[b]{0.1\textwidth}
\includegraphics[width=\textwidth]{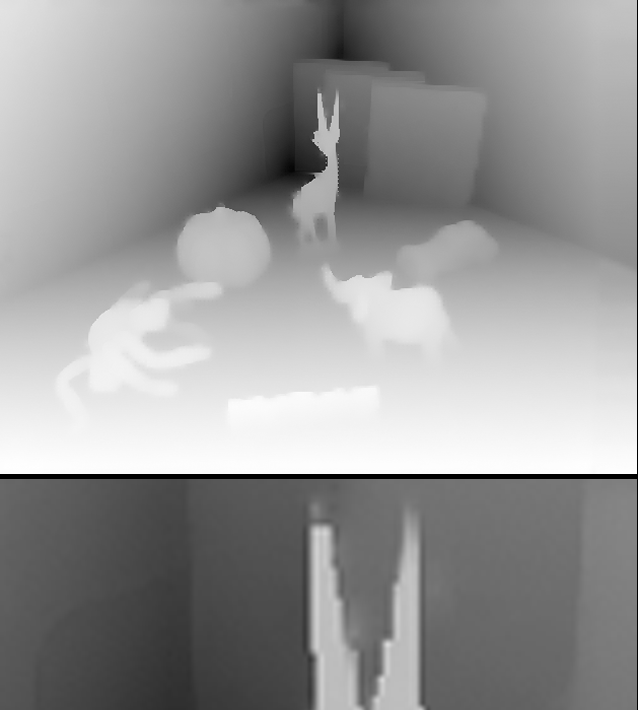}\\
\includegraphics[width=\textwidth]{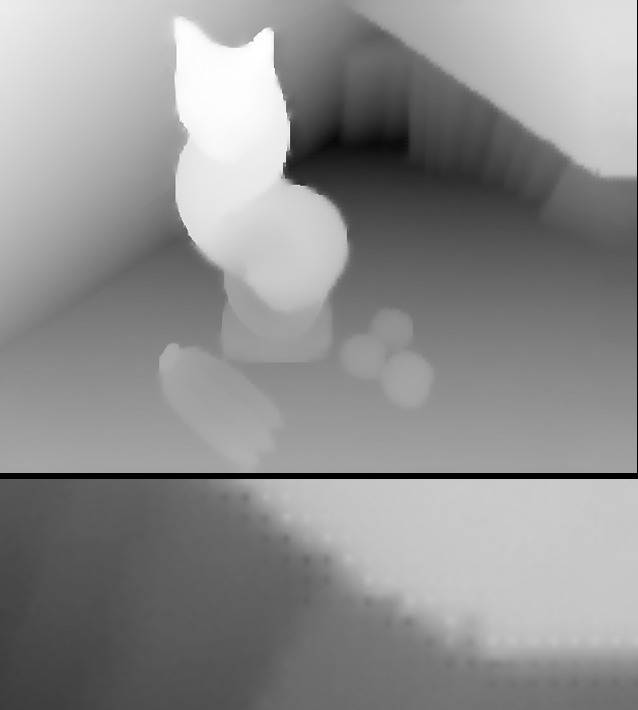}\\
\includegraphics[width=\textwidth]{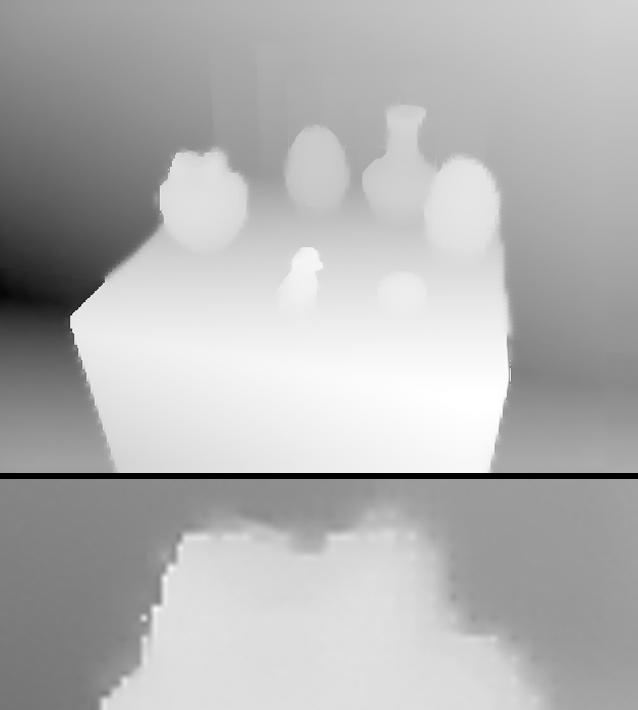}
\end{minipage}
}
\subfigure[Ours]{
\begin{minipage}[b]{0.1\textwidth}
\includegraphics[width=\textwidth]{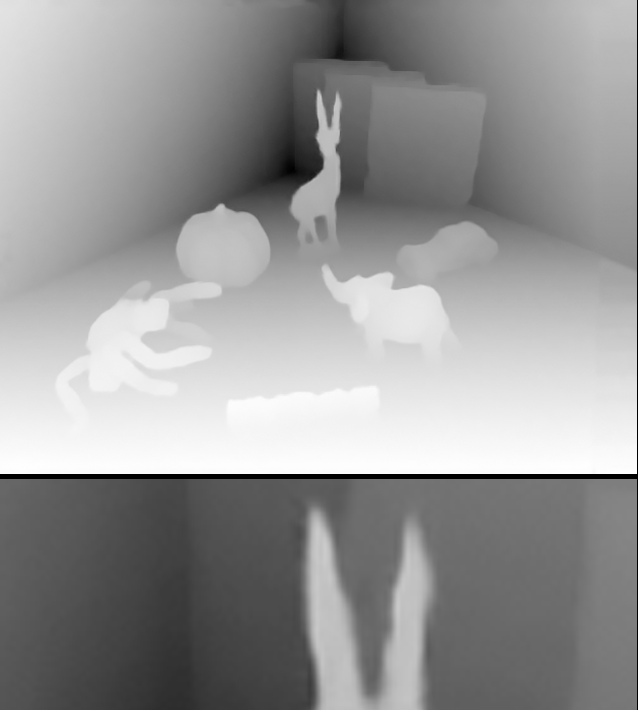}\\
\includegraphics[width=\textwidth]{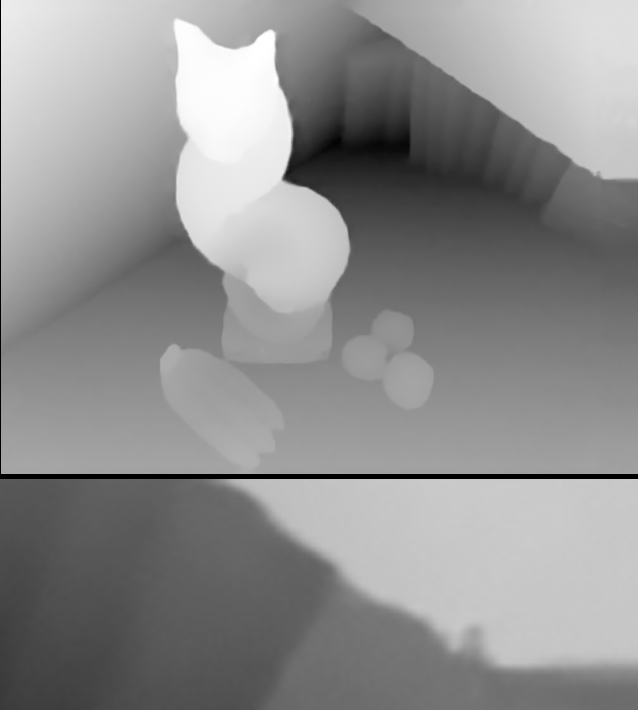}\\
\includegraphics[width=\textwidth]{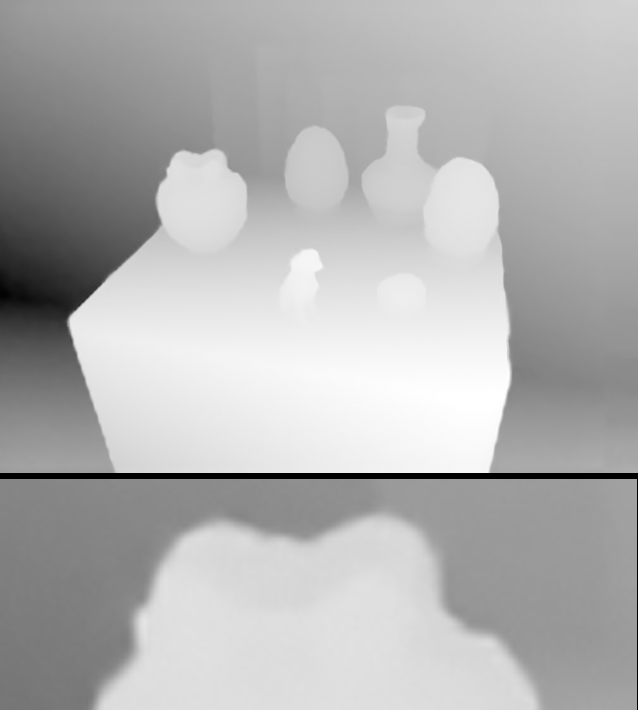}
\end{minipage}
}
\caption{Visual comparisons of depth map SR results(4$\times$).}
\label{Figure-DepthVisualComparison}
\end{figure*}

\begin{table*}[t]
\centering
\caption{Average RMSE comparison of depth map SR on the four benchmark datesets. We map depth values into the range[0,255] for each data set. Bold Values indicate the best performance and underscored values indicate the second best.} \label{tab:RMSE-DepthSR}
\setlength{\tabcolsep}{1.5mm}{
\begin{tabular}{|c|c c c c|c c c c|c c c c|c c c c|}
  \hline
  Methods & \multicolumn{4}{c|}{Middlebury~\cite{Scharstein2007Learning,Hirschmuller2007Evaluation}} & \multicolumn{4}{c|}{Lu~\cite{Lu2014Depth}} & \multicolumn{4}{c|}{NYU~\cite{Silberman2012Indoor}} & \multicolumn{4}{c|}{SUN~\cite{Song2015SUN}}\\
  \hline
  {}& 2$\times$ & 4$\times$ & 8$\times$ & 16$\times$ &
      2$\times$ & 4$\times$ & 8$\times$ & 16$\times$ &
      2$\times$ & 4$\times$ & 8$\times$ & 16$\times$ &
      2$\times$ & 4$\times$ & 8$\times$ & 16$\times$ \\
  \hline

  Bicubic & 1.97 & 4.45 & 7.59 & 11.88  &
            2.16 & 5.08 & 9.22 & 14.27  &
            2.11 & 4.51 & 7.85 & 12.36  &
            2.44 & 3.89 & 5.62 & 8.05   \\

  Kopf~\cite{Kopf2007Joint}     & 1.41  &  2.44  &  3.82  &  \underline{6.12}    &
            1.39 & 2.99 & 5.06 & \underline{7.51}   &
            \underline{1.22} & 2.30 & 4.60 & 7.23   &
            \textbf{1.74} & \textbf{2.73} & \underline{3.85} & \underline{5.72}   \\

  He~\cite{He2010Guided}      & 2.43   &  4.10  &  7.26  &  11.71    &
            3.03 & 4.89 & 8.85 & 14.10  &
            2.31 & 4.10 & 7.53 & 12.20  &
            2.43 & 3.56 & 5.40 & 7.96   \\

  Li~\cite{Li2016Deep}      & \underline{1.28}   &  \underline{2.14}  &  \underline{3.63}  &  6.13    &
            \underline{1.24} & \underline{2.55} & \underline{4.71} & 7.66   &
            1.30 & \underline{2.00} & \underline{3.41} & \underline{5.67}   &
            \underline{1.75} & 3.25 & 5.04 & 7.56   \\

  Ham~\cite{Ham2015Robust}      & 1.91   &  3.15  &  5.01  &  8.81    &
            2.40 & 4.61 & 7.50 & 11.49   &
            1.63 & 3.01 & 7.20 & 11.00   &
            2.13 & 3.35 & 5.19 & 7.75   \\


  Ours    & \textbf{1.16} & \textbf{1.79} & \textbf{3.22} & \textbf{5.54}   &
            \textbf{0.83} & \textbf{1.86} & \textbf{4.04} & \textbf{6.28}   &
            \textbf{0.93} & \textbf{1.59} & \textbf{3.17} & \textbf{5.09}   &
            1.81 &\underline{2.75} & \textbf{3.64} & \textbf{5.16}   \\

  \hline
\end{tabular}}
\end{table*}

\begin{figure*}[t]
\centering
\subfigure[Scribbles]{
\label{Figure.sub.1}
\includegraphics[width=0.15\textwidth]{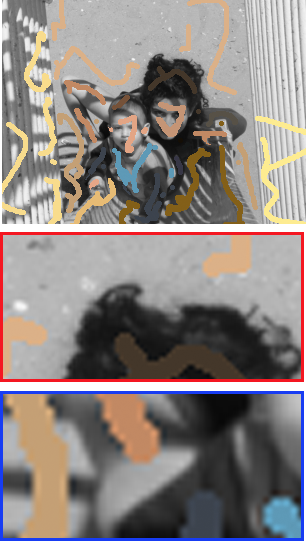}}
\subfigure[GT~\cite{Levin2004Colorization}]{
\label{Figure.sub.2}
\includegraphics[width=0.15\textwidth]{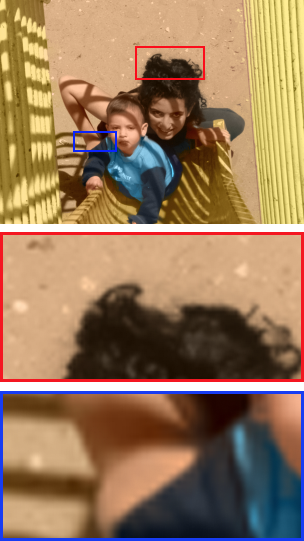}}
\subfigure[Bicubic]{
\label{Figure.sub.3}
\includegraphics[width=0.15\textwidth]{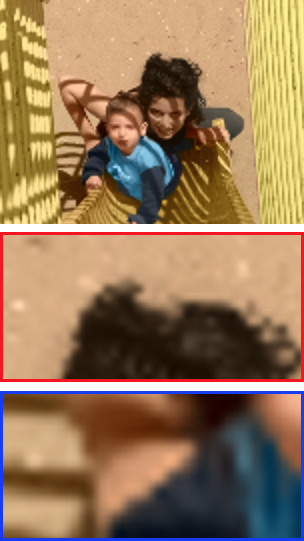}}
\subfigure[He~\cite{He2010Guided}]{
\label{Figure.sub.4}
\includegraphics[width=0.15\textwidth]{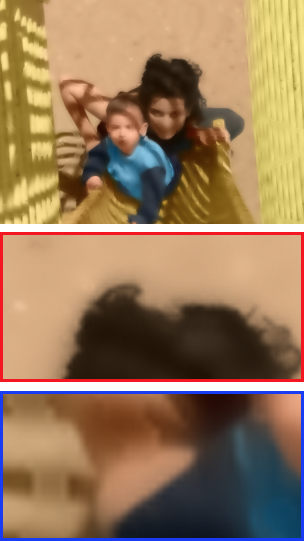}}
\subfigure[Li~\cite{Li2016Deep}]{
\label{Figure.sub.5}
\includegraphics[width=0.15\textwidth]{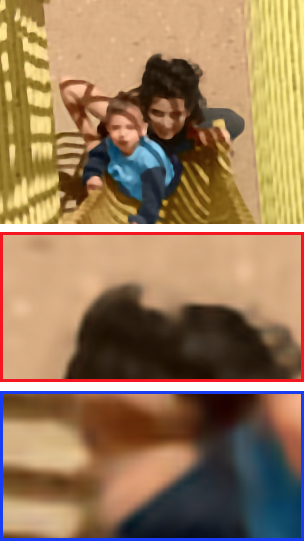}}
\subfigure[Ours]{
\label{Figure.sub.6}
\includegraphics[width=0.15\textwidth]{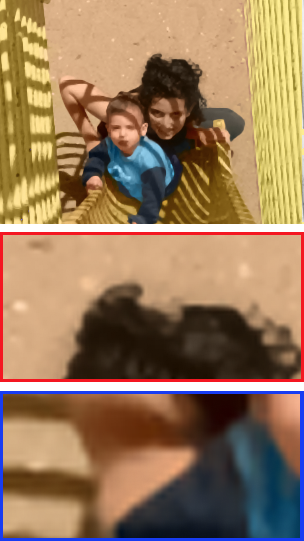}}
\caption{Visual comparisons of chromaticity map SR results(2$\times$).}
\label{Figure.VisualChromaticitySR}
\end{figure*}

\begin{figure*}[h]
\centering
\subfigure[Input(guidance)]{
\label{Figure.sub.1}
\includegraphics[width=0.15\textwidth]{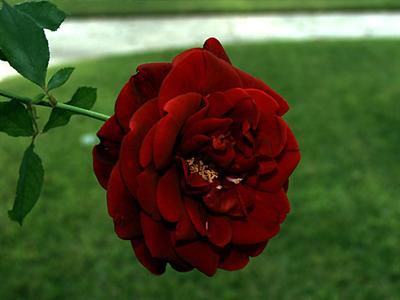}}
\subfigure[GT]{
\label{Figure.sub.2}
\includegraphics[width=0.15\textwidth]{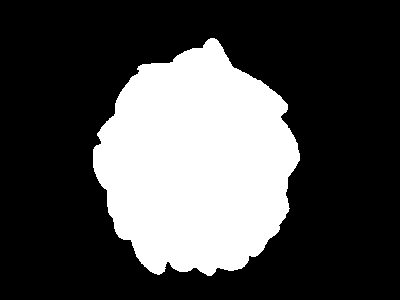}}
\subfigure[Bicubic]{
\label{Figure.sub.3}
\includegraphics[width=0.15\textwidth]{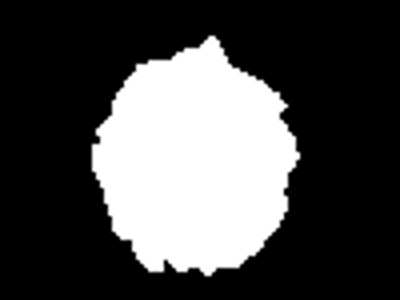}}
\subfigure[He~\cite{He2010Guided}]{
\label{Figure.sub.4}
\includegraphics[width=0.15\textwidth]{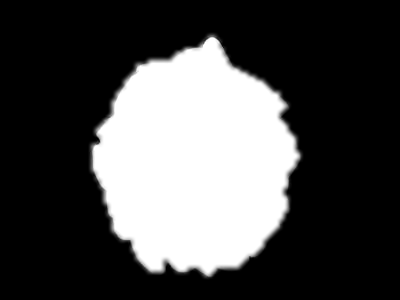}}
\subfigure[Li~\cite{Li2016Deep}]{
\label{Figure.sub.5}
\includegraphics[width=0.15\textwidth]{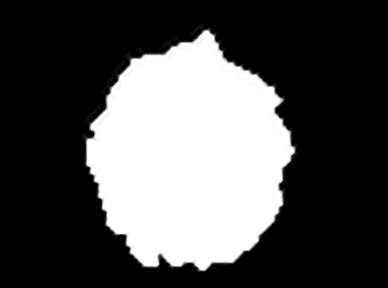}}
\subfigure[Ours]{
\label{Figure.sub.6}
\includegraphics[width=0.15\textwidth]{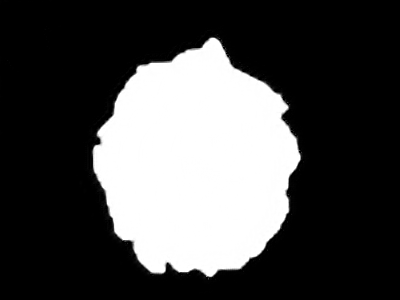}}
\caption{Visual comparisons of saliency map SR results(4$\times$).}
\label{Figure.VisualSaliencySR}
\end{figure*}

\subsection{Depth Map SR}
We compare our JCNP model with several state-of-art methods, including local-based methods~\cite{Kopf2007Joint,He2010Guided}, global-based method~\cite{Ham2015Robust}, and CNN-based method~\cite{Li2016Deep}. We first provide the visual comparisons with several state-of-art SR methods in Figure~\ref{Figure-DepthVisualComparison}. It is observed that the results of He~\cite{He2010Guided} and Li~\cite{Li2016Deep} are over smoothed (Figure~\ref{Figure-DepthVisualComparison}(d)(e)). The JBU~\cite{Kopf2007Joint} and Ham~\cite{Ham2015Robust} approaches transfer erroneous details (Figure~\ref{Figure-DepthVisualComparison}(f)(g)). In contrast, the SR depth maps reconstructed by ours JCNP model are sharper and has less artifacts than the other methods (Figure~\ref{Figure-DepthVisualComparison}(h)).

We also provide a numerical comparison with the state-of-art SR methods using the average RMSEs in Table~\ref{tab:RMSE-DepthSR}. The four benchmark test data sets include:
(1) 30 RGB/depth hole-filled pairs from~Middlebury data sets~\cite{Scharstein2007Learning,Hirschmuller2007Evaluation}, (2) Lu data sets~\cite{Lu2014Depth} contains 6 RGB/depth pairs captured by ASUS Xtion Pro, (3) the 449 RGB/D testing pairs in NYU v2 data sets~\cite{Silberman2012Indoor}, (4) 299 RGB-D image pairs from align\_kv2 sub-folder obtained by the Kinect V2 sensor in SUN RGB/D data sets~\cite{Song2015SUN}.

The RMSE values are computed from the code packages with suggested parameters provided by the authors.
As shown in the Table~\ref{tab:RMSE-DepthSR}, our JCNP model achieves the state-of-the-art performance, especially for Middlebury~\cite{Scharstein2007Learning,Hirschmuller2007Evaluation}, Lu~\cite{Lu2014Depth} and NYU~\cite{Silberman2012Indoor} data sets. For the Sun data set, our results are comparable to the best results in the $2\times$ and $4\times$ SR case. Thanks to the large recept field provided by our JCNP model, our results outperform the best results in $8\times$ and $16\times$ SR.

\subsection{Chromaticity Map SR}

\begin{table}[t]
\begin{center}
\caption{Average RMSE comparisons of different methods for chromaticity map SR and saliency map SR.} \label{tab:NumericalChromaticitySR}
\setlength{\tabcolsep}{2mm}{
\begin{tabular}{|c|c|c|c|c|}
  \hline
  Methods & Bicubic & He\cite{He2010Guided} & Li\cite{Li2016Deep} & Ours
  \\
  \hline
  2$\times$ Chromaticity  & 10.91 & 11.18 & 9.72 & \textbf{6.73} \\
    4$\times$ Chromaticity  & 20.54 & 17.99 & 14.18 & \textbf{13.41} \\
   \hline
  2$\times$ Saliency    & 14.08  & 14.99 & 12.29 & \textbf{11.19} \\
  4$\times$ Saliency  &  23.05 & 21.08  & 18.27 & \textbf{15.46} \\
  \hline
\end{tabular}}
\end{center}
\end{table}

We apply our model to chromaticity map SR. The test data are six color-scribbles/colorized image pairs provided by the authors of \cite{Levin2004Colorization}. We first compute the LR map by down-sampling the colorized image and use it as the target image. We then use the HR color-scribbles map as guidance image to construct the output HR Chromaticity map.

 Figure~\ref{Figure.VisualChromaticitySR} shows the visual comparisons with the method of Bicubic, He~\cite{He2010Guided} and Li~\cite{Li2016Deep}. Our results show more faithful edges and less color bleeding artifacts (\ref{Figure.VisualChromaticitySR}(f)). We also provide the qualitative evaluations in Table~\ref{tab:NumericalChromaticitySR}. We use the colorized image as the ground truth and calculate the RMSE values. The RMSE values in Table~\ref{tab:NumericalChromaticitySR} show that our results best approximate the ground truth.

\subsection{Saliency Map SR}
We also apply our trained JCNP model in saliency map SR. We random collect eight color/saliency map pairs from MSRA10K salient object database~\cite{13iccv/Cheng_Saliency} in the test. We use nearest-neighbor down-sampling to get the LR saliency map and use HR color map as the guidance image. We compare our results with those of Bicubic, He~\cite{He2010Guided} and Li~\cite{Li2016Deep}.
Visual comparison in Figure~\ref{Figure.VisualSaliencySR} and numerical comparison in~\ref{tab:NumericalChromaticitySR} show that our results can better reconstruct image content by better transferring useful structure from the guidance image to the target image.

\section{Conclusion and Future Work}
We have proposed a joint convolution neural pyramid for joint image SR. Our model can enable large receptive fields and transfer useful structures from the guidance image to enhance the resolution of the target image. Experimental results verified the performance of method on joint depth map SR, chromaticity map SR and saliency map SR. In the future, we would like to apply our model on more applications. Also, another future work is to study the impact of different loss functions on our model.


\begin{thebibliography}{10}

\bibitem{Scharstein2002A}
Scharstein, Daniel, Szeliski, and Richard,
\newblock ``A taxonomy and evaluation of dense two-frame stereo correspondence
  algorithms,''
\newblock {\em IJCV}, vol. 47, no. 1-3, pp. 7--42, 2002.

\bibitem{Kopf2007Joint}
Johannes Kopf, Michael~F. Cohen, Dani Lischinski, and Matt Uyttendaele,
\newblock ``Joint bilateral upsampling,''
\newblock in {\em SIGGRAPH}, 2007, p.~96.

\bibitem{He2010Guided}
Kaiming He, Jian Sun, and Xiaoou Tang,
\newblock ``Guided image filtering,''
\newblock in {\em ECCV}, 2010, pp. 1--14.

\bibitem{Ham2015Robust}
Bumsub Ham, Minsu Cho, and Jean Ponce,
\newblock ``Robust image filtering using joint static and dynamic guidance,''
\newblock in {\em CVPR}, 2015, pp. 4823--4831.

\bibitem{Hui2016Depth}
Tak~Wai Hui, Change~Loy Chen, and Xiaoou Tang,
\newblock ``Depth map super-resolution by deep multi-scale guidance,''
\newblock in {\em ECCV}, 2016, pp. 353--369.

\bibitem{Li2016Deep}
Yijun Li, Jia~Bin Huang, Narendra Ahuja, and Ming~Hsuan Yang,
\newblock ``Deep joint image filtering,''
\newblock in {\em ECCV}, 2016, pp. 154--169.

\bibitem{DBLP:journals/corr/ShenCTJ17}
Xiaoyong Shen, Ying{-}Cong Chen, Xin Tao, and Jiaya Jia,
\newblock ``Convolutional neural pyramid for image processing,''
\newblock {\em CoRR}, vol. abs/1704.02071, 2017.

\bibitem{Yang2007Spatial}
Qingxiong Yang, Ruigang Yang, James Davis, and David Nister,
\newblock ``Spatial-depth super resolution for range images,''
\newblock in {\em CVPR}, 2007, pp. 1--8.

\bibitem{Liu2013Joint}
Ming~Yu Liu, Oncel Tuzel, and Yuichi Taguchi,
\newblock ``Joint geodesic upsampling of depth images,''
\newblock in {\em CVPR}, 2013, pp. 169--176.

\bibitem{Lu2015Sparse}
Jiajun Lu and David Forsyth,
\newblock ``Sparse depth super resolution,''
\newblock in {\em CVPR}, 2015, pp. 2245--2253.

\bibitem{Diebel2005An}
James Diebel and Sebastian Thrun,
\newblock ``An application of markov random fields to range sensing,''
\newblock {\em NIPS}, pp. 291--298, 2005.

\bibitem{Park2011High}
Jaesik Park, Hyeongwoo Kim, Yu~Wing Tai, Michael~S Brown, and Inso Kweon,
\newblock ``High quality depth map upsampling for 3d-tof cameras,''
\newblock in {\em ICCV}, 2011, pp. 1623--1630.

\bibitem{Aodha2012Patch}
Oisin~Mac Aodha, Neill D.~F. Campbell, Arun Nair, and Gabriel~J. Brostow,
\newblock ``Patch based synthesis for single depth image super-resolution,''
\newblock in {\em ECCV}, 2012, pp. 71--84.

\bibitem{Ferstl2013Image}
David Ferstl, Christian Reinbacher, Rene Ranftl, Matthias Ruether, and Horst
  Bischof,
\newblock ``Image guided depth upsampling using anisotropic total generalized
  variation,''
\newblock in {\em ICCV}, 2013, pp. 993--1000.

\bibitem{Yang2014Color}
J.~Yang, X.~Ye, K.~Li, C.~Hou, and Y.~Wang,
\newblock ``Color-guided depth recovery from rgb-d data using an adaptive
  autoregressive model,''
\newblock {\em TIP}, vol. 23, no. 8, pp. 3443--3458, 2014.

\bibitem{Lei2017Depth}
Jianjun Lei, Lele Li, Huanjing Yue, Wu~Feng, Nam Ling, and Chunping Hou,
\newblock ``Depth map super-resolution considering view synthesis quality,''
\newblock {\em TIP}, vol. 26, no. 4, pp. 1732, 2017.

\bibitem{Krizhevsky2012ImageNet}
Alex Krizhevsky, Ilya Sutskever, and Geoffrey~E. Hinton,
\newblock ``Imagenet classification with deep convolutional neural networks,''
\newblock in {\em NIPS}, 2012, pp. 1097--1105.

\bibitem{Dong2014Learning}
Chao Dong, Change~Loy Chen, Kaiming He, and Xiaoou Tang,
\newblock ``Learning a deep convolutional network for image super-resolution,''
\newblock in {\em ECCV}, 2014, pp. 184--199.

\bibitem{Ren2017Image}
Haoyu Ren, Mostafa Elkhamy, and Jungwon Lee,
\newblock ``Image super resolution based on fusing multiple convolution neural
  networks,''
\newblock in {\em CVPR}, 2017, pp. 1050--1057.

\bibitem{Huang2016Densely}
Gao Huang, Zhuang Liu, and Kilian~Q. Weinberger,
\newblock ``Densely connected convolutional networks,''
\newblock in {\em CVPR}, 2017.

\bibitem{Silberman2012Indoor}
Nathan Silberman, Derek Hoiem, Pushmeet Kohli, and Rob Fergus,
\newblock ``Indoor segmentation and support inference from rgbd images,''
\newblock in {\em ECCV}, 2012, pp. 746--760.

\bibitem{Lu2014Depth}
Si~Lu, Xiaofeng Ren, and Feng Liu,
\newblock ``Depth enhancement via low-rank matrix completion,''
\newblock in {\em CVPR}, 2014, pp. 3390--3397.

\bibitem{Scharstein2007Learning}
D.~Scharstein and C.~Pal,
\newblock ``Learning conditional random fields for stereo,''
\newblock in {\em CVPR}, 2007, pp. 1--8.

\bibitem{Hirschmuller2007Evaluation}
Heiko Hirschmuller and Daniel Scharstein,
\newblock ``Evaluation of cost functions for stereo matching,''
\newblock in {\em CVPR}, 2007, pp. 1--8.

\bibitem{Song2015SUN}
Shuran Song, Samuel~P. Lichtenberg, and Jianxiong Xiao,
\newblock ``Sun rgb-d: A rgb-d scene understanding benchmark suite,''
\newblock in {\em CVPR}, 2015, pp. 567--576.

\bibitem{Levin2004Colorization}
Anat Levin, Dani Lischinski, and Yair Weiss,
\newblock ``Colorization using optimization,''
\newblock in {\em SIGGRAPH}, 2004, pp. 689--694.

\bibitem{13iccv/Cheng_Saliency}
Ming-Ming Cheng, Jonathan Warrell, Wen-Yan Lin, Shuai Zheng, Vibhav Vineet, and
  Nigel Crook,
\newblock ``Efficient salient region detection with soft image abstraction,''
\newblock in {\em ICCV}, 2013, pp. 1529--1536.

\end{thebibliography}

\end{document}